\documentclass[conference]{IEEEtran}
\IEEEoverridecommandlockouts
 
\usepackage{graphicx}
\usepackage{amsmath}
\usepackage{url}
\usepackage{xcolor}
\usepackage{enumitem}
\usepackage{listings}
\usepackage{algorithm}
\usepackage{algpseudocode}
\usepackage[most,listings]{tcolorbox}
\usepackage{multirow}
\usepackage{makecell}
\usepackage{tikz}
\usepackage{tabularray}
\usepackage{booktabs}
\usepackage{nicefrac}
\usepackage{amssymb}
\usepackage{marvosym} % \Letter (envelope) for the corresponding-author mark
\newcommand{\eqmark}{\textsuperscript{\dag}}      % co-first author
\newcommand{\cormark}{\textsuperscript{\Letter}}  % corresponding author

\definecolor{rkup}{RGB}{0,150,0}\definecolor{rkdn}{RGB}{200,0,0}

\newcommand{\rankdn}[1]{\textcolor{rkdn}{$-$#1}}

\usetikzlibrary{fit,positioning,arrows.meta}
\usepackage[most,listings]{tcolorbox}
\usepackage[font=small,labelfont=bf]{caption}

\definecolor{MotivBg}{RGB}{247,248,250}
\definecolor{MotivFrame}{RGB}{180,180,180}
\definecolor{LineNumGray}{RGB}{140,140,140}
\definecolor{SnagRed}{RGB}{170,50,50}
\definecolor{bggray}{RGB}{245,245,245}
\definecolor{framegray}{RGB}{180,180,180}

\setlist[itemize]{leftmargin=*, topsep=1pt, itemsep=1pt}
\setlist[enumerate]{leftmargin=*, topsep=1pt, itemsep=1pt}

\lstdefinestyle{motivation}{
  language=Python,
  basicstyle=\ttfamily\footnotesize,
  numbers=left,
  numberstyle=\tiny\color{LineNumGray},
  numbersep=6pt,
  frame=none,
  tabsize=2,
  breaklines=true,
  showstringspaces=false,
  commentstyle=\itshape\color{gray!70}
}

\newtcblisting{motivationbox}{
  listing only,
  listing options={style=motivation},
  colback=MotivBg,
  colframe=MotivFrame,
  arc=3pt,
  boxrule=0.6pt,
  left=12pt,
  right=6pt,
  top=4pt,
  bottom=4pt,
}

\def\BibTeX{{\rm B\kern-.05em{\sc i\kern-.025em b}\kern-.08em
    T\kern-.1667em\lower.7ex\hbox{E}\kern-.125emX}}
\begin{document}
\title{EarlyEval: Cheaper Agent Evaluation via Early Outcome Prediction}

\author{
\IEEEauthorblockN{
Yuling Shi\textsuperscript{1}\eqmark\thanks{\eqmark~Equal contribution.}\thanks{\cormark~Corresponding author.},
Zhensu Sun\textsuperscript{2}\eqmark,
Junsen Dong\textsuperscript{1},
Chengcheng Wan\textsuperscript{3,4},
David Lo\textsuperscript{2},
Xiaodong Gu\textsuperscript{1}\cormark
}
\IEEEauthorblockA{
\textsuperscript{1}\textit{Shanghai Jiao Tong University}, Shanghai, China \quad
\textsuperscript{2}\textit{Singapore Management University}, Singapore\\
\textsuperscript{3}\textit{East China Normal University}, Shanghai, China \quad
\textsuperscript{4}\textit{Shanghai Innovation Institute}, Shanghai, China\\
\{yuling.shi, 171263615, xiaodong.gu\}@sjtu.edu.cn \quad
\{zssun, davidlo\}@smu.edu.sg \quad
ccwan@sei.ecnu.edu.cn
}
}

\maketitle

\begin{abstract}
Evaluating LLM agents is essential for guiding their development, yet it has grown prohibitively expensive: a single pass of a frontier model over an agentic benchmark can cost hundreds to thousands of dollars, a price paid repeatedly across iterative development cycles. 
Prior efforts, centered on benchmark distillation, reduce the number of evaluation tasks but leave the cost of executing each retained task untouched.
In this work, we introduce early outcome prediction, a complementary axis of efficiency that instead cuts cost within each task.
Our key insight is that an agent's final outcome is often evident from its intermediate behavior well before execution completes.
We instantiate this idea in EarlyEval, a lightweight framework that trains a pair of LightGBM success and failure classifiers over behavioral, textual, and reference-solution features, and halts an agent run the moment either classifier crosses a calibrated confidence threshold, adding negligible per-step overhead. Across three benchmarks, SWE-bench Verified, TerminalBench, and Toolathlon, EarlyEval can eliminate 13\%--26\% of agent steps and up to 44.1\% input tokens and 29.4\% output tokens at 89\%--97\% prediction accuracy, while perturbing per-agent resolve rates by only one to two percentage points on average. 
\end{abstract}

\section{Introduction}\label{sec:intro}
Evaluation is fundamental to the development of LLM agents~\cite{yao2023reactsynergizingreasoningacting,liu2025agentbenchevaluatingllmsagents,DBLP:conf/acl/PengSWZSG26}. It is not only a final verdict on whether a system performs well, but also a compass that guides the design of new ones, telling researchers and developers which changes help and which hurt~\cite{liu2025agentbenchevaluatingllmsagents,kapoor2024aiagentsmatter,DBLP:journals/corr/abs-2607-01211,DBLP:conf/emnlp/FangSSWG25}. In practice this guidance is consumed repeatedly: a single development cycle may involve dozens of iterative benchmark runs as a team tunes prompts, adjusts scaffolding, and finetunes model variants~\cite{kapoor2024aiagentsmatter}.

However, the cost of running an agentic benchmark has risen sharply~\cite{kapoor2024aiagentsmatter,Per23}. On SWE-bench Verified~\cite{jimenez2024swebench}, a single evaluation pass of a frontier model costs several hundred dollars, and benchmarks with longer rollouts run several times higher, reaching into the thousands of dollars per pass~\cite{openhands_index}. Costs of this magnitude pose a real resource barrier for agent development, putting frequent evaluation out of reach for many practitioners and slowing the iteration loop that drives progress~\cite{kapoor2024aiagentsmatter}.

To mitigate these costs, prior work has predominantly focused on benchmark distillation~\cite{Per23}, which downsizes a benchmark into a smaller subset of representative tasks. Common approaches include selecting a handful of anchor tasks~\cite{Viv23} or constructing a compact proxy test set whose scores closely track those of the full suite~\cite{Pol24}. While effective, this line of work exclusively reduces the number of tasks within a benchmark~\cite{Per23,Viv23,Pol24}, leaving the per-task execution cost untouched. Consequently, the remaining tasks that must be retained remain as computationally and financially expensive to execute as before~\cite{kapoor2024aiagentsmatter}.

\begin{figure}[t]
\centering
\includegraphics[width=\linewidth]{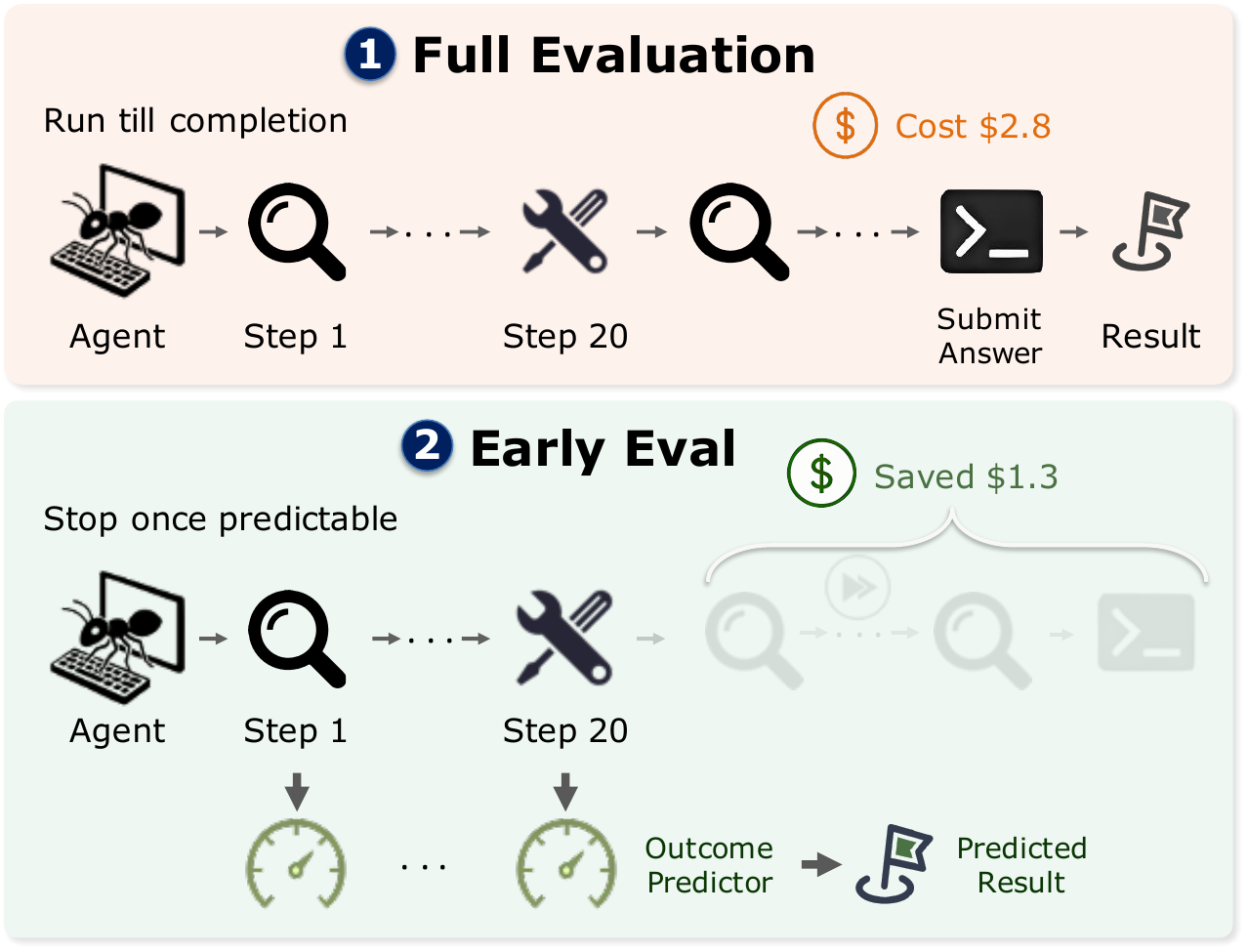}
\caption{Comparison between full evaluation and EarlyEval.}
\vspace{-0.3cm}
\label{fig:motivation}
\end{figure}

In this work, we approach the problem from a different angle.
It is driven by a key insight: for most evaluation tasks, an agent does not need to run to completion for its final score to be accurately inferred (Figure~\ref{fig:motivation}).
The ultimate outcome is often highly predictable from the agent’s intermediate behavior long before the final answer is generated. Sometimes it is legible against a reference solution, i.e., the moment an agent applies the correct one-line edit, task resolution can be confidently anticipated, rendering subsequent steps such as running the test suite redundant.
But such signals are frequently intrinsic to the trajectory itself, requiring no reference: an agent that repeatedly retries the same edit against an unchanging error message has, in effect, already announced its eventual failure.
We exploit both, and, as our ablations show, lean primarily on these reference-free behavioral signals, allowing it to operate even on benchmarks that release no gold solutions.

Grounded in this observation, we propose EarlyEval, an effective paradigm that infers an agent's final evaluation outcome as soon as its behavioral trajectory exhibits a strong indicator of success or failure. Crucially, these indicators can be learned from the historical behavior of other agents on the same benchmark. Such historical trajectories are readily available in practice: agentic benchmarks are typically released alongside various runs, and their public leaderboards accumulate large pools of outcome-labeled submissions over time.
Specifically, we collect the full trajectories of a diverse set of agents on the target benchmark, labeled with their ground-truth outcomes, and train two classifiers over the behavior accumulated up to any given step: a success classifier that triggers when a task can be confidently declared resolved, and a failure classifier that triggers when failure becomes highly certain. To evaluate a new agent, we apply both classifiers at each step of its execution. The moment either classifier crosses a predefined confidence threshold, we terminate the run and record the predicted outcome; if both remain below the threshold, the agent is permitted to proceed. The thresholds are fixed in advance and expose a tunable knob that trades off how early we stop against how reliably the predicted outcome matches the true one.

To evaluate the effectiveness of EarlyEval, we conduct experiments on three agentic benchmarks: SWE-bench Verified~\cite{jimenez2024swebench}, TerminalBench~\cite{merrill2026terminalbenchbenchmarkingagentshard}, and Toolathlon~\cite{li2026tooldecathlonbenchmarkinglanguage}, which span software issue resolution, shell automation, and tool use. Across these benchmarks we collect more than $21{,}000$ outcome-labeled trajectories from $16$, $37$, and $22$ distinct agents, where each agent is a scaffolding harness paired with a base LLM. To ensure EarlyEval judges an agent it has never seen, we adopt a leave-one-agent-out protocol that holds out one agent at a time and trains only on the remaining ones. The experimental results show that, on SWE-bench Verified, EarlyEval can halt roughly $35\%$ of runs at $95\%$ prediction accuracy, eliminating $26\%$ of execution steps along with $33\%$ of input and $29\%$ of output tokens, while shifting each agent's measured resolve rate by only $1.1$ percentage points on average. Furthermore, this early-stopped evaluation can largely reproduce the ranking of the full-run leaderboard, attaining a Spearman rank correlation of $\rho=0.991$ over all 16 agents, with only three adjacently ranked agents shifting position by a single rank. The same conclusions hold on TerminalBench and Toolathlon: even under the strict leakage controls, EarlyEval saves $13\%$ to $25\%$ of execution steps at $89\%$ to $97\%$ accuracy, keeps the average resolve-rate deviation within roughly two percentage points, and preserves rankings at $\rho \geq 0.959$. 

In summary, we make the following contributions:
\begin{itemize}
\item We propose the concept of early outcome prediction, a new dimension of evaluation efficiency that reduces computational costs within individual tasks. By terminating an agent's rollout the moment its outcome becomes predictable, this approach complements, rather than replaces, existing benchmark distillation methods.
\item We present EarlyEval, a lightweight and plug-and-play framework that trains LightGBM~\cite{NIPS2017_6449f44a} success and failure classifiers over a rich feature space spanning behavioral trajectories, textual context, and reference-solution metadata. Coupled with a calibrated threshold-based halting rule, EarlyEval introduces negligible per-step inference overhead.
\item We conduct rigorous leave-one-agent-out evaluations across three diverse agentic benchmarks: SWE-bench Verified, TerminalBench, and Toolathlon. The results demonstrate that EarlyEval substantially reduces execution steps and token consumption while robustly preserving original per-agent resolve rates and overall leaderboard rankings.
\end{itemize}

\section{Background and Motivation}\label{sec:bg}

\subsection{Agentic Benchmarks are Expensive}
Modern agentic benchmarks evaluate a model by letting it act over many steps on each task: reading files, running commands, calling tools, and revising its approach when something fails. This multi-step rollout is what makes such benchmarks faithful to real use, and it is also what makes them costly. Every step issues at least one model call, and a single task can run for dozens of steps before it terminates, so the token bill for one task dwarfs that of a conventional question-answering item. The cost then compounds across the full task set.

To quantify these costs, we draw on the OpenHands Index~\cite{openhands_index}, a public leaderboard that records the measured dollar cost of running recent models through the OpenHands agent~\cite{wang2025openhands} on five software-engineering benchmarks: SWE-bench Verified~\cite{jimenez2024swebench}, SWT-bench~\cite{mundler2024swtbench}, Commit0~\cite{zhao2024commit0}, GAIA~\cite{mialon2024gaia}, and SWE-bench Multimodal~\cite{yang2025swebenchmm}. Table~\ref{tab:cost} reports the cost of one evaluation pass for three frontier models. Even SWE-bench Verified, which is among the less expensive benchmarks on a per-task basis, amounts to several hundred dollars for a single pass. 
Benchmarks with longer rollouts are considerably more expensive: a single pass over SWE-bench Multimodal reaches into the thousands of dollars, exceeding \$2,200 for the most costly model in Table~\ref{tab:cost} and surpassing \$1,000 for two of the three models.

All of these figures correspond to a single evaluation of a single agent configuration. In practice, a team tuning an agent re-evaluates after each modification to the prompt, the scaffold, or the underlying model, and repeats the process for every baseline under comparison. A development cycle involving dozens of such runs multiplies a few hundred dollars per pass into a substantial total, placing frequent evaluation beyond the reach of many practitioners.

\begin{table}[t]
\centering
\caption{Cost of one evaluation pass on the benchmarks tracked by the OpenHands Index, using the OpenHands agent. Figures retrieved June 2026.}
\label{tab:cost}
\setlength{\tabcolsep}{3pt}
\begin{tabular}{lcccc}
\toprule
\multirow{2}{*}{\textbf{Benchmark}} & \multirow{2}{*}{\textbf{\#Tasks}} & \multicolumn{3}{c}{\textbf{Full-run Cost (USD)}} \\
\cmidrule(lr){3-5}
 & & \textbf{Claude 5} & \textbf{GPT-5.5} & \textbf{Gemini 3.1 Pro} \\
\midrule
SWE-bench Verified   & 500 & \$715     & \$760     & \$935 \\
SWT-bench            & 500 & \$735     & \$460     & \$810 \\
Commit0              & 54  & \$674     & \$300     & \$64  \\
GAIA                 & 165 & \$1{,}305 & \$122     & \$297 \\
SWE-bench Multimodal & 517 & \$2{,}270 & \$1{,}453 & \$641 \\
\bottomrule
\end{tabular}
\end{table}

\subsection{Outcomes are Often Foreseeable Early}

For many tasks, the final outcome is discernible from an agent's behavior well
before the run reaches its end. We make this concrete through an observer that
monitors a trajectory with access to the ground-truth solution for each task,
enabling it to compare the agent's intermediate state against the correct
answer at any point. To see what this observer would conclude, consider a
publicly released OpenHands~\cite{wang2025openhands} trajectory (tianocore\_\_edk2-pytool-library-372) from a real
issue in \texttt{tianocore/edk2-pytool-library}, where a path utility
documented to return a forward-slash relative path instead returned a path with
backslashes.
The run spans 45 steps and terminates with a patch that resolves the task, yet
the substantive work concludes well before the final step.

By step~20 the agent has written a script that reproduces the bug. At
step~23 it makes its sole modification to the source code, normalizing the
path separators in a single line. After that, the agent makes no further
changes to the source, though it continues testing in various directions.
Having observed the correct fix applied at step~23, the observer can already
conclude that the task is resolved. Stopping there would record the identical
evaluation outcome at roughly half the cost.

\begin{table*}[t]
\centering
\caption{Features EarlyEval extracts from a partial run $\tau_{:k}$, grouped into
three families.}
\label{tab:features}
\footnotesize
\begin{tblr}{
colspec = {@{}Q[l,m] Q[l,m] Q[r,m] p{0.62\linewidth}@{}},
  rowsep  = 3pt,
  row{1}  = {font=\bfseries},
  hline{1,Z} = {0.08em},
  hline{2,7,10} = {0.05em},
}
Family & Feature group & \# & Description \\
\SetCell[r=5]{m} Behavioral
 & Activity counts      & 37  & Cumulative counts through step $k$ at two granularities: overall volume (steps, actions, observations, tool calls, distinct tools, and the character length of the action, feedback, and task text) and per-category occurrences spanning file views and searches, the various edit operations (create, replace, insert, undo), test, Python, and CLI executions, git operations, and submissions. \\
 & Last step            & 11  & Properties of the most recent step, including its action category and sub-type, the number of tools it called, whether it produced any output, and whether its feedback signaled a tool error, traceback, or test pass/fail. \\
 & Event timing         & 18  & The temporal structure of key events (edit, test, code run, submission, error, traceback, file read): the step at which each first occurs, boolean flags marking whether each has occurred at all, and the number of steps elapsed since each last occurred. \\
 & Working pattern      & 32  & Derived signals characterizing how the agent works: rhythm ratios (reads per edit, edits per test, bash-vs-editor balance, error and submission rates, mean text length), stalling and risky-control-flow indicators (repeated actions, searches, or views, no-edit and consecutive-read streak lengths, submitting without testing or editing after submission), and the volume, pacing, and action-overlap of its reasoning and assistant-message text. \\
 & Error \& test status & 17  & Indicators for whether each error type (traceback, assertion, type, value, syntax, import, file-not-found, timeout, permission) and each test outcome (pass, fail, all-passed) has been observed, alongside the latest and best failure counts and whether the failure count is trending down. \\
\SetCell[r=3]{m} Textual
 & Task prompt          & 64  & SVD embedding of the TF-IDF representation of the task (issue) description. \\
 & Action text          & 128 & SVD embeddings of the TF-IDF representations of the full action history and of the most recent action. \\
 & Feedback text        & 128 & SVD embeddings of the TF-IDF representations of all environment feedback and of the most recent feedback. \\
\SetCell[r=2]{m} Reference
 & Gold descriptors     & 28  & Attributes of the reference solution: patch size in characters, lines, and hunks, the number of files changed, fail-to-pass and pass-to-pass test counts, API, import, and exception token counts, directory depth, and the repository, difficulty, and version. \\
 & Prefix--gold overlap & 54  & Jaccard overlap and hit counts between the files, API symbols, and test names the agent has touched and those appearing in the reference solution, capturing how far the run has progressed toward the gold fix. \\
\end{tblr}
\end{table*}

\section{Approach}\label{sec:method}

\begin{figure*}[t]
\centering
\includegraphics[width=\linewidth]{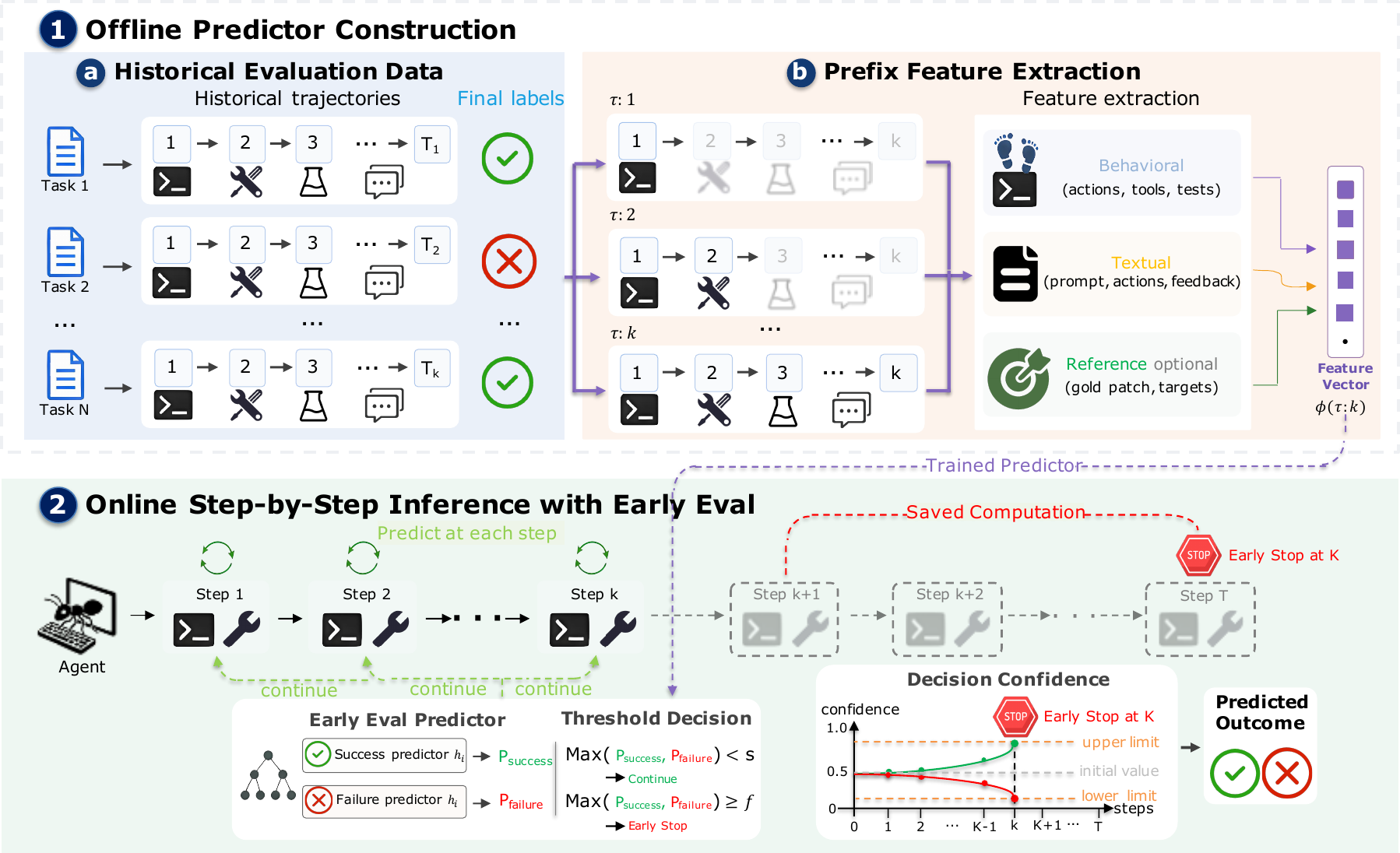}
\caption{Overview of EarlyEval: (1) offline predictor construction, which extracts prefix features from historical trajectories and their final labels, and (2) online step-by-step inference, which halts the run once confidence in the predicted outcome is sufficiently high.}
\label{fig:overview}
\end{figure*}
\subsection{Problem Definition}
Early outcome prediction is the problem of inferring an agent's final score
on a benchmark task from its partial run, before the run reaches completion,
so that the remaining steps need not be executed.

Specifically, we consider an agent $\mathcal{A}$ running on a task $t$ drawn from a
benchmark $\mathcal{B}$. The agent produces a trajectory
$\tau = (e_1, e_2, \ldots, e_T)$, where each event $e_k$ records the action
taken at step $k$ together with the resulting observation, and $T$ is the
total number of steps until the agent halts. At termination, the benchmark
assigns a binary score $y \in \{0, 1\}$ indicating failure or success.
Obtaining $y$ in the conventional way requires the agent to complete all $T$
steps.

An early-outcome predictor monitors the run as it unfolds and may, at any
step $k < T$, issue a prediction $\hat{y} \in \{0, 1\}$ and halt the run. If
it does not yet have sufficient confidence, it lets the agent continue to the
next step. When the predictor fires, its output is recorded as the task's
score in place of the true outcome $y$.

\subsection{Overview}
EarlyEval predicts an agent's final outcome on a benchmark task from a partial
trajectory and halts the execution as soon as the eventual outcome becomes statistically evident. 
An overview of this sequential inference workflow is illustrated in Stage~2 of Figure~\ref{fig:overview}.
Given a specific task, the agent interacts with the environment step by step, generating an evolving trajectory. At each step, EarlyEval extracts multi-modal features from the accumulated partial trajectory.
These features are then fed into the EarlyEval predictor, which computes success and failure confidences over the final outcome.
The framework employs a dual-threshold decision mechanism based on these confidences:
if either confidence reaches or exceeds its designated threshold, execution is immediately intercepted to output a Predicted Outcome;
otherwise, the agent is allowed to persist in its execution until either a threshold is breached or the run completes and its full-run outcome is recorded.

To support this early halting capability, the underlying predictors are trained on historic agent runs that have already been evaluated on the benchmark. 
As shown in Stage~1 of Figure~\ref{fig:overview}, each historical submission supplies a complete trajectory paired with its ground-truth outcome, providing the supervision signal exploited by our framework. We expand every trajectory into a sequence of labeled prefixes and train two distinct classifiers over them: a \emph{success predictor} that triggers when the current prefix provides sufficient evidence of task success, and a \emph{failure predictor} that triggers when it indicates inevitable failure. 

\subsection{Processing Training Data}\label{subsec:data}
For a given benchmark $\mathcal{B}$, we collect a pool of agent trajectories
$\{\mathcal{A}_1,\dots,\mathcal{A}_M\}$ evaluated across the tasks within
$\mathcal{B}$. Each trajectory $\tau=(e_1,\dots,e_T)$ is associated with a binary
evaluation score $y\in\{0,1\}$ assigned by the benchmark upon execution termination, where $y=1$ denotes success and $y=0$ denotes failure. Trajectories shorter than 10 steps are discarded, as they rarely contain sufficient signal for meaningful optimization.

Stage~1 of Figure~\ref{fig:overview} formalizes the pipeline for constructing our training data. We cross-reference the text of benchmark tasks with their historical trajectories. These trajectories are decomposed into constituent prefixes to derive multi-modal features, while the corresponding final labels are directly mapped to supervisory targets.

Specifically, for a trajectory of length $T$, we construct prefixes $\tau_{:k}=(e_1,\dots,e_k)$ for $k=0,1,\dots,T$, and pair each prefix with the trajectory's \emph{final} outcome label $y$ (the final labels in Fig.~\ref{fig:overview}). Each prefix is then mapped to a fixed-length feature vector $\phi(\tau_{:k})\in\mathbb{R}^d$. As illustrated in Figure~\ref{fig:overview} and detailed in Table~\ref{tab:features}, the coordinates of $\phi$ capture multi-modal readings of the prefix, which are clustered into three distinct families:

\begin{itemize}
\item \textbf{Behavioral Features} capture run progression invariants across different tasks. These include volume and pacing metrics, the structural composition of the immediate step, milestone execution timing, error and test signals extracted from environment feedback, and behavioral patterns indicating agent stalling or premature submission.

\item \textbf{Textual Features} encode the natural-language context of the trajectory. 
Textual data is isolated into distinct semantic blocks: the task prompt (1 block), the full action history and the most recent action (2 blocks), and all environment feedback alongside the most recent feedback (2 blocks). Each individual block is vectorized independently using TF-IDF over word $n$-grams and subsequently compressed to $64$ dimensions via Truncated Singular Value Decomposition (SVD) prior to concatenation, resulting in a 64-dimensional embedding for the prompt and 128-dimensional embeddings for the action and feedback groups respectively. Vectorizing blocks independently preserves their semantic boundaries, while the SVD reduction maintains the aggregate textual dimensionality in the low hundreds, ensuring per-step inference remains computationally inexpensive.

\item \textbf{Reference-Solution Features} are leveraged when the benchmark provides ground-truth human patches (e.g., SWE-bench Verified), instantiating the oracle-observer intuition outlined in Section~\ref{sec:bg}. Beyond encoding the properties of the gold patch itself, these features measure the agent's convergence toward the reference solution by computing structural overlaps between the files, symbols, and tests present in the current prefix and those in the gold solution. Benchmarks lacking released reference patches omit this feature family and rely exclusively on behavioral and textual features.
\end{itemize}

\subsection{Training the Prediction Model}\label{subsec:model}
For a given benchmark, EarlyEval trains a single pair of agent-agnostic predictors, which can be directly deployed for any unseen new agent.
Specifically, the predictors judge partial trajectories using gradient-boosted decision tree ensembles trained via LightGBM over the feature representation $\phi(\tau_{:k})$.
We select this architecture because tree ensembles can evaluate a several-hundred-dimensional feature vector in well under a millisecond on a single CPU core. This efficiency allows EarlyEval to re-score the trajectory at every step with negligible computational overhead. By contrast, an LLM-based judge would incur substantial inference costs at each execution step, effectively offsetting the execution compute our framework aims to conserve.

EarlyEval optimizes two separate ensembles over the representation $\phi$: a \emph{success predictor} $h_{+}$ and a \emph{failure predictor} $h_{-}$. Although both models ingest the same feature vector, they are optimized against inverted target sets. For a prefix derived from a trajectory with a final outcome $y$, the success predictor targets $y=1$, whereas the failure predictor targets $1-y=1$ (i.e., $y=0$). Consequently, $h_{+}$ specializes in recognizing trajectories converging toward success, while $h_{-}$ isolates patterns indicative of impending failure. Training two predictors rather than a single joint classifier allows positive and negative evidence to accumulate independently, reflecting the empirical reality that success and failure are signaled by fundamentally asymmetric behaviors. Furthermore, it creates an explicit unconfident region—where both predictors output low probabilities—allowing the agent to continue execution when the final outcome remains ambiguous (corresponding to the Continue branch in Fig.~\ref{fig:overview}).

To prevent data leakage, we partition the trajectory pool by task into a training fold and a held-out validation fold; all prefixes originating from a given trajectory are strictly restricted to the same side of the split. The training fold is used to fit the parameters of both predictors, while the validation fold is reserved for probability calibration (Section~\ref{subsec:inference}).

To prevent prolonged trajectories from dominating the optimization loss, we weight each prefix instance by $1/(T+1)$, thereby ensuring that every trajectory contributes identical total mass to the objective function regardless of its length. Both predictors share a unified LightGBM hyperparameter configuration (including the number of leaves, learning rate, and boosting rounds), which is tuned once on the validation fold and held constant across all benchmarks.

\subsection{Predicting Early Outcomes}\label{subsec:inference}
At each step of an unobserved agent run, we extract the feature vector $\phi$ from the accumulated partial trajectory and input it into both ensembles. Because regularized, weight-balanced tree ensembles typically distort output probability scales, we recalibrate the raw scores using Platt scaling. Specifically, a one-dimensional logistic regression maps the raw ensemble score $\hat{s}$ to a calibrated probability:
$$p=\sigma\!\big(a\,\operatorname{logit}(\hat{s})+b\big)$$
where the scalar parameters $a,b$ are fitted on the held-out validation split under the per-prefix sample weights defined in Section~\ref{subsec:model}. A distinct calibrator is optimized for each predictor within each cross-validation fold. Because this transformation is monotonic, calibration preserves each predictor's sample ranking and resulting AUC; its sole function is to rescale outputs so that confidence thresholds carry a consistent, comparable meaning across both predictors and evaluation folds.

The calibrated probabilities $p_{+}$ and $p_{-}$ parameterize the threshold-based decision rule illustrated in Figure~\ref{fig:overview} to determine whether the run can be halted with sufficient confidence. We compare $p_{+}$ against a success threshold $s$ and $p_{-}$ against a failure threshold $f$. As conceptualized by the Threshold Decision logic in Fig.~\ref{fig:overview}, a run is stopped and marked with a Predicted Outcome (either success or failure) at the first step where $p_{+}\!\ge s$ or $p_{-}\!\ge f$. In the rare event that both thresholds are breached simultaneously on the same step, the earlier crossing chronologically takes precedence. While both probabilities remain below their respective thresholds (i.e., $p_{+}\!<s$ and $p_{-}\!<f$), the system defers commitment and allows the agent to proceed to subsequent steps.
The thresholds $s$ and $f$ thus dictate the stringency of evidence EarlyEval demands before intervention; they can be set based on the target accuracy-efficiency trade-off, where higher thresholds yield superior prediction accuracy at the expense of deferred termination and reduced compute savings.
\section{Experimental Setup}\label{sec:setup}
We evaluate EarlyEval to systematically assess its predictive accuracy and compute efficiency. In this section, we introduce our benchmark suites, evaluated agent architectures, performance metrics, and implementation details. The rationale behind our setup is driven by answering our four core research questions:

\begin{itemize}
\item \textbf{RQ1:} How accurately does EarlyEval predict the evaluation outcome of an unseen agent, and how much computation is thereby saved by stopping them early?

\item \textbf{RQ2:} To what extent are the ground-truth downstream agent rankings preserved under EarlyEval's early stopping?

\item \textbf{RQ3:} How robust is EarlyEval to the omission of specific features?

\item \textbf{RQ4:} How does the LightGBM predictor compare against alternative architectures in the cost--fidelity trade-off?
\end{itemize}

\subsection{Benchmarks and Collected Trajectories}
We evaluate EarlyEval across three distinct agentic benchmarks that require agents to interact over multi-step environment trajectories.
In our framework, an agent is defined as a specific prompting or scaffolding harness paired with an underlying LLM.
Each unique agent configuration attempts the full task pool of its respective benchmark.

\begin{itemize}
    \item \textbf{SWE-bench Verified} comprises $500$ human-validated GitHub issues sourced across $12$ Python repositories. Each issue ships with a gold patch and a fail-to-pass/pass-to-pass test suite, so an agent is scored \emph{resolved} when its final patch passes the suite. We collect trajectories from a single scaffold, mini-SWE-agent, paired with $16$ base LLMs spanning the Claude, GPT-5, Gemini, GLM, DeepSeek, Devstral, Kimi, and MiniMax families, for $7{,}805$ trajectories in total.

    \item \textbf{TerminalBench} targets shell interaction, comprising 89 command-line automation tasks.
    Our collected pool for this benchmark contains $37$ distinct agent configurations and $6{,}757$ trajectories, where each configuration is run with multiple rollouts per task.
    The scaffolds include mini-SWE-agent, the provider-native CLI agents (Codex, Claude Code, and Gemini CLI), OpenHands, and Terminus-2; the base models span GPT-5, GPT-5-mini, Claude-Haiku-4.5, Claude-Opus-4.5, and Gemini-2.5-Pro.

    \item \textbf{Toolathlon} is a suite for complex API and tool-use patterns, with $108$ tasks. Every run uses the native Toolathlon scaffold. We collected $22$ base LLMs with three rollouts per task, yielding $7{,}116$ trajectories.
\end{itemize}

Notably, neither TerminalBench nor Toolathlon releases per-task reference solutions.
Thus, on these two benchmarks, EarlyEval disables the Reference-Solution features and relies exclusively on behavioral and textual features.

\subsection{Evaluation Protocol}
In a realistic deployment setting, EarlyEval must judge an unseen agent configuration, either an unseen model or an unseen scaffold.
Therefore, we enforce a rigorous \textbf{leave-one-agent-out} evaluation protocol.
We rotate through the agent pool, holding out one test agent at a time.
The outcome predictors are trained solely on the trajectories of the remaining agents, and the held-out agent's test runs are scored using these trained predictors.
Metrics are calculated per held-out fold and aggregated.
For TerminalBench, the collected trajectories are heterogeneous: for a held-out agent, both its model and its scaffold may individually appear in the training data, though never combined into the same agent.
To rule out this residual leakage, we evaluate TerminalBench under two complementary settings: (1) \emph{no same model in training} and (2) \emph{no same scaffold in training}.
Each setting removes from the training pool every trajectory that shares the held-out agent's model (setting 1) or scaffold (setting 2) before fitting that agent's predictors.

\subsection{Evaluation Metrics} \label{subsec:metrics}
We assess EarlyEval along three complementary axes:

\begin{itemize}
\item \textbf{Decision quality}:
We treat each halted trajectory as a binary prediction of its final outcome and compare it against $y$.
\textbf{Accuracy} is the fraction of halted trajectories whose predicted label matches the ground-truth outcome.
Because each predictor commits only on the trajectories it actually halts, we additionally report the \textbf{precision} of each predictor in isolation: among the trajectories the success (resp.\ failure) predictor halts and labels $y=1$ (resp.\ $y=0$), the fraction whose ground-truth outcome matches.
Precision thus captures the reliability of a predictor's early-stop commitments, which is the quantity of interest when the success and failure heads are evaluated separately.

\item \textbf{Compute efficiency}:
We quantify the resources eliminated by early termination relative to running every trajectory to completion.
\textbf{Coverage} is the proportion of trajectories that are halted early (i.e., for which a predictor crosses its threshold, $s$ or $f$, before natural completion), and therefore upper-bounds the attainable savings.
$\Delta$\textbf{Steps} is the relative change in executed environment-interaction steps, $\Delta\text{Steps} = \nicefrac{\sum S_\text{early}}{\sum S_\text{full}} - 1$, where $S_\text{early}$ and $S_\text{full}$ denote step counts under early stopping and full execution; a negative value is the fraction of steps eliminated. 
Analogously, $\Delta$\textbf{Token}\textsubscript{in} and $\Delta$\textbf{Token}\textsubscript{out} report the corresponding relative changes in input (prompt) and output (generation) tokens, capturing the compounding context-window cost that step counts alone do not reflect.

\item \textbf{Evaluation fidelity}:
Finally, we measure how faithfully early-stopped evaluation reproduces the conclusions of full execution, in both absolute scores and relative comparisons.
For a single agent, $\Delta$\textbf{Pass@1} is the signed difference, in percentage points, between its early-stopped resolve rate and its ground-truth resolve rate, so a positive value means early stopping over-credits the agent and a negative value means it under-credits.
Aggregating over the agent pool, $\Delta|$\textbf{Pass@1}$|$ is the mean of the per-agent \emph{absolute} deviations, which quantifies the typical distortion of the headline performance number irrespective of sign.
To assess whether relative comparisons survive, we report \textbf{Spearman's rank correlation} $\rho$ between the agent ranking induced by EarlyEval and the full-run ranking, together with the \textbf{rank shift} $\Delta$\textbf{Rank}, the change in an agent's ordinal position between the two rankings.
\end{itemize}

\subsection{Implementation Details}
We implement EarlyEval in Python on top of scikit-learn and LightGBM. Prior to the SVD projection, each textual block is vectorized over word unigrams and bigrams with a minimum document frequency of $5$ and a vocabulary capped at $30{,}000$ terms. Both gradient-boosted predictors share a regularized LightGBM configuration: learning rate $0.03$, $31$ leaves, maximum depth $6$, minimum child samples $200$, row and feature subsampling of $0.75$ and $0.70$, and $\ell_1/\ell_2$ regularization weights of $0.5$ and $10.0$, trained for up to $2{,}000$ boosting rounds with early stopping after $50$ rounds without validation improvement.
Within each fold we reserve $15\%$ of the training trajectories as the validation split that drives early stopping and Platt calibration.
All experiments use a fixed random seed $42$.
\section{Results}
\begin{table*}[t]\centering
\small
\caption{RQ1: Performance and computational savings of EarlyEval across different decision thresholds and benchmarks under the leave-one-agent-out evaluation protocol.}
\label{tab:rq1}
\scriptsize
\begin{tblr}{
  colspec = {Q[c,wd=0.8cm] *{4}{Q[c,wd=1.05cm]} Q[wd=5pt] *{4}{Q[c,wd=1.05cm]} Q[wd=5pt] *{4}{Q[c,wd=1.15cm]}},
  colsep  = 0pt,
  rowsep  = 1.3pt,
  row{1}  = {font=\bfseries},
  row{2}  = {font=\bfseries},
  hline{1} = {0.08em},
  hline{2} = {2-5,7-10,12-15}{0.05em},
  hline{3} = {0.05em},
  hline{10,17,24} = {0.05em},
  hline{Z} = {0.08em},
  cell{1}{2}  = {c=4}{c},
  cell{1}{7}  = {c=4}{c},
  cell{1}{12} = {c=4}{c},
  cell{3}{1}  = {c=15}{l},
  cell{10}{1} = {c=15}{l},
  cell{17}{1} = {c=15}{l},
  cell{24}{1} = {c=15}{l},
}
 & Success-only & & & & & Failure-only & & & & & Dual & & & \\
Thre. & Prec. & Cov. & $\Delta$Steps & $\Delta|$Pass@1$|$ & & Prec. & Cov. & $\Delta$Steps & $\Delta|$Pass@1$|$ & & $\Delta$Steps & $\Delta$Token\textsubscript{in} & $\Delta$Token\textsubscript{out} & $\Delta|$Pass@1$|$ \\
\textit{SWE-bench Verified} & & & & & & & & & & & & & & \\
0.75 & 88.3\% & 50.0\% & -33.1\% & 5.8\% &  & 87.4\% & 28.9\% & -30.3\% & 3.6\% &  & -63.4\% & -81.5\% & -69.7\% & 4.1\% \\
0.80 & 89.6\% & 46.9\% & -30.4\% & 4.9\% &  & 89.0\% & 26.9\% & -28.2\% & 3.0\% &  & -58.6\% & -75.7\% & -64.2\% & 3.5\% \\
0.85 & 90.9\% & 42.1\% & -26.3\% & 3.8\% &  & 90.8\% & 24.2\% & -25.5\% & 2.2\% &  & -51.8\% & -67.6\% & -56.8\% & 2.9\% \\
0.90 & 92.1\% & 35.7\% & -21.1\% & 2.8\% &  & 93.9\% & 20.5\% & -21.6\% & 1.3\% &  & -42.7\% & -54.8\% & -46.9\% & 2.3\% \\
\textbf{0.95} & \textbf{93.9\%} & \textbf{20.4\%} & \textbf{-10.6\%} & \textbf{1.3\%} &  & \textbf{96.7\%} & \textbf{14.4\%} & \textbf{-15.4\%} & \textbf{0.5\%} &  & \textbf{-26.0\%} & \textbf{-32.7\%} & \textbf{-28.7\%} & \textbf{1.1\%} \\
0.97 & 93.5\% & 10.0\% & -5.0\% & 0.7\% &  & 98.3\% & 10.4\% & -11.6\% & 0.2\% &  & -16.6\% & -22.0\% & -18.0\% & 0.6\% \\
\textit{TerminalBench: no same model in training} & & & & & & & & & & & & & & \\
0.75 & 72.3\% & 7.7\% & -6.0\% & 2.2\% &  & 84.8\% & 36.8\% & -44.3\% & 5.6\% &  & -50.3\% & -75.8\% & -52.3\% & 4.4\% \\
0.80 & 77.3\% & 5.4\% & -3.8\% & 1.2\% &  & 86.5\% & 32.4\% & -39.6\% & 4.4\% &  & -43.4\% & -67.8\% & -45.9\% & 4.0\% \\
0.85 & 81.8\% & 3.4\% & -2.1\% & 0.6\% &  & 88.0\% & 26.8\% & -33.3\% & 3.2\% &  & -35.4\% & -57.0\% & -38.0\% & 3.2\% \\
\textbf{0.90} & \textbf{82.7\%} & \textbf{1.6\%} & \textbf{-0.8\%} & \textbf{0.3\%} &  & \textbf{89.4\%} & \textbf{19.3\%} & \textbf{-24.6\%} & \textbf{2.0\%} &  & \textbf{-25.4\%} & \textbf{-42.7\%} & \textbf{-27.9\%} & \textbf{2.1\%} \\
0.95 & 83.8\% & 0.2\% & 0.0\% & 0.0\% &  & 92.6\% & 8.4\% & -10.7\% & 0.6\% &  & -10.7\% & -19.6\% & -12.6\% & 0.7\% \\
0.97 & -- & 0.0\% & 0.0\% & 0.0\% &  & 95.8\% & 3.6\% & -4.7\% & 0.2\% &  & -4.7\% & -8.6\% & -5.6\% & 0.2\% \\
\textit{TerminalBench: no same scaffold in training} & & & & & & & & & & & & & & \\
0.75 & 61.4\% & 14.6\% & -13.3\% & 5.6\% &  & 87.0\% & 23.4\% & -26.6\% & 3.1\% &  & -39.8\% & -58.9\% & -39.1\% & 4.9\% \\
0.80 & 65.5\% & 10.3\% & -8.9\% & 3.6\% &  & 88.8\% & 17.5\% & -19.9\% & 2.0\% &  & -28.9\% & -44.8\% & -28.6\% & 3.6\% \\
\textbf{0.85} & \textbf{69.0\%} & \textbf{6.2\%} & \textbf{-5.0\%} & \textbf{1.9\%} &  & \textbf{91.8\%} & \textbf{11.2\%} & \textbf{-12.7\%} & \textbf{0.9\%} &  & \textbf{-17.7\%} & \textbf{-29.2\%} & \textbf{-17.4\%} & \textbf{2.0\%} \\
0.90 & 77.0\% & 2.8\% & -2.1\% & 0.7\% &  & 93.3\% & 5.1\% & -5.0\% & 0.4\% &  & -7.0\% & -11.5\% & -6.6\% & 0.8\% \\
0.95 & 65.6\% & 0.6\% & -0.5\% & 0.2\% &  & 92.7\% & 0.8\% & -0.8\% & 0.1\% &  & -1.3\% & -2.2\% & -1.1\% & 0.2\% \\
0.97 & 36.8\% & 0.1\% & 0.0\% & 0.1\% &  & 100.0\% & 0.2\% & -0.2\% & 0.0\% &  & -0.2\% & -0.2\% & -0.1\% & 0.1\% \\
\textit{Toolathlon} & & & & & & & & & & & & & & \\
0.75 & 81.2\% & 1.6\% & -0.8\% & 0.3\% &  & 89.6\% & 50.0\% & -41.9\% & 5.2\% &  & -42.7\% & -68.2\% & -50.6\% & 5.0\% \\
0.80 & 80.7\% & 0.8\% & -0.3\% & 0.2\% &  & 91.9\% & 44.1\% & -36.9\% & 3.6\% &  & -37.2\% & -62.7\% & -45.0\% & 3.5\% \\
0.85 & 81.8\% & 0.3\% & -0.1\% & 0.1\% &  & 93.9\% & 36.6\% & -30.6\% & 2.2\% &  & -30.7\% & -55.6\% & -38.0\% & 2.3\% \\
\textbf{0.90} & \textbf{--} & \textbf{0.0\%} & \textbf{0.0\%} & \textbf{0.0\%} &  & \textbf{96.6\%} & \textbf{27.6\%} & \textbf{-23.0\%} & \textbf{0.9\%} &  & \textbf{-23.0\%} & \textbf{-44.1\%} & \textbf{-29.4\%} & \textbf{0.9\%} \\
0.95 & -- & 0.0\% & 0.0\% & 0.0\% &  & 98.6\% & 16.3\% & -13.4\% & 0.2\% &  & -13.4\% & -28.3\% & -17.7\% & 0.2\% \\
0.97 & -- & 0.0\% & 0.0\% & 0.0\% &  & 99.4\% & 9.4\% & -7.5\% & 0.1\% &  & -7.5\% & -15.1\% & -10.1\% & 0.1\% \\
\end{tblr}
\vspace{-0.2cm}
\end{table*}
\subsection{RQ1: Prediction Accuracy and Compute Savings}

We first evaluate the accuracy with which EarlyEval predicts the evaluation outcomes of unseen agents, alongside the computational savings achieved by prematurely halting trajectories. Adhering to a leave-one-agent-out cross-validation protocol, we sweep the decision thresholds across $\{0.75, 0.80, 0.85, 0.90, 0.95, 0.97\}$ and report performance under three configurations: the success predictor alone (Success-only), the failure predictor alone (Failure-only), and the full dual-threshold mechanism (Dual). For each benchmark, we identify an optimal operating point that balances aggressive computational reduction against fidelity to the original evaluation metric. TerminalBench results are reported under two leakage-controlled settings, where neither the identical model nor the identical scaffold is present in the training fold.
Specifically, the recommended operating point for each benchmark is selected as the lowest threshold where the dual-mechanism absolute deviation ($\Delta|\text{Pass@1}|$) remains within approximately 2 percentage points, representing the most aggressive configuration that preserves the headline metric.

EarlyEval yields substantial computational savings while maintaining the integrity of the benchmark metrics. Specifically, it reduces agent execution steps by $26.0\%$ on SWE-bench Verified at a $0.95$ threshold, and by $23.0\%$ on Toolathlon at a $0.90$ threshold. Concurrently, the absolute deviation in the task resolution rate remains within approximately one percentage point for both benchmarks ($\Delta|\text{Pass@1}|$ of $1.1\%$ and $0.9\%$, respectively).
Token-level reductions are even more pronounced because early termination eliminates the compounding cost of expanding context windows; at the aforementioned SWE-bench threshold, input and output token volumes decrease by $32.7\%$ and $28.7\%$, respectively.
Lowering the decision thresholds systematically trades metric fidelity for enhanced computational savings. For instance, reducing the threshold from $0.95$ to $0.75$ on SWE-bench Verified elevates the dual step reduction from $26.0\%$ to $63.4\%$, while the $\Delta|\text{Pass@1}|$ distortion increases from $1.1\%$ to $4.1\%$.
This monotonic trade-off demonstrates that thresholds can be tuned in practice to satisfy the specific accuracy-efficiency constraints of any given evaluation budget.

The success predictor exhibits its highest reliability on SWE-bench Verified, where its precision consistently ranges between $88.3\%$ and $93.9\%$ across all thresholds. On other benchmarks, however, its performance diminishes significantly; precision drops to $61.4\%$--$69.0\%$ on TerminalBench (under the no-same-scaffold split), and coverage collapses toward zero on Toolathlon, leaving the corresponding columns virtually empty at higher thresholds.
In contrast, the failure predictor demonstrates robust precision across all settings, reaching $96.7\%$ on SWE-bench, $89.4\%$--$96.6\%$ on TerminalBench, and $96.6\%$--$99.4\%$ on Toolathlon at the designated operating points.
This performance asymmetry indicates that while the default dual configuration provides acceptable baseline performance, practitioners deploying EarlyEval can further optimize efficiency by validating individual predictors against historical trajectories, as a benchmark-specific calibration can identify the more trustworthy predictor and yield a superior operating point.
Moreover, at nearly every operating point the dual step reduction equals the sum of its success-only and failure-only counterparts (e.g., 
$-10.6\%+(-15.4\%)=-26.0\%$ on SWE-bench Verified at $0.95$). Since a run 
halts at the first threshold crossing, this additivity implies that the two 
predictors almost never fire on the same trajectory, i.e., positive and negative evidence rarely coincide within a single run.

Finally, the two TerminalBench configurations reveal that scaffold behavior is inherently more challenging to model than base LLM behavior. Withholding the test agent's scaffold degrades the success predictor more severely than withholding its base model, causing peak precision to drop from $82.7\%$ (no same model) to $69.0\%$ (no same scaffold), while the attainable dual step reduction at the recommended operating point shrinks from $25.4\%$ to $17.7\%$. Because a scaffold dictates the structural rhythm of a trajectory—orchestrating how actions, environmental feedback, and milestones are sequenced—an unseen scaffold fundamentally perturbs the behavioral features that EarlyEval relies upon, whereas an unseen model leaves this structural skeleton comparatively stable.

\begin{tcolorbox}[size=title]
{\textbf{Answer to RQ1:}} EarlyEval eliminates $29.2\%$--$44.1\%$ of input tokens and $17.4\%$--$29.4\%$ of output tokens while bounding the absolute resolve-rate deviation within $2.1$ percentage points ($\Delta|\text{Pass@1}|$). This demonstrates that substantial computational savings are attainable without materially distorting downstream benchmark metrics.
\end{tcolorbox}

\begin{table}[t]\centering
\caption{RQ2: Downstream leaderboard fidelity and agent ranking preservation under EarlyEval at benchmark-specific optimal operating points.}
\label{tab:rq2}
\scriptsize
\begin{tblr}{
  colspec   = {Q[l,wd=4cm] Q[c,wd=1cm] Q[c,wd=1cm] Q[c,wd=1cm]},
  colsep    = 3pt,
  rowsep    = 1.2pt,
  row{1}    = {font=\bfseries},
  row{6,11,16,21} = {font=\itshape},
  hline{1}  = {0.08em},
  hline{2}  = {0.05em},
  hline{7,12,17} = {0.05em},
  hline{Z}  = {0.08em},
  cell{2}{1}  = {c=4}{l},
  cell{7}{1}  = {c=4}{l},
  cell{12}{1} = {c=4}{l},
  cell{17}{1} = {c=4}{l},
}
Agent & $\Delta$Pass@1 & $\Delta$Rank & $\Delta$Steps \\
\textit{SWE-bench Verified (threshold = 0.95)} & & & \\
Gemini-3-Pro      & $-1.4$ & 0 & -23.8\% \\
GPT-5.2-High      & $-0.4$ & 0 & -14.7\% \\
Claude-Sonnet-4.5 & $-0.2$ & 0 & -16.2\% \\
All 16 ($\rho\,0.991$, 81\% unchanged) & $+0.8$ & -- & -26.0\% \\
\textit{TerminalBench (threshold = 0.90): no same model in training} & & & \\
Terminus-2 + Claude-Opus-4.5  & $-3.6$ & 0 & -23.9\% \\
Claude-Code + Claude-Opus-4.5 & $-1.2$ & \rankdn{2} & -20.3\% \\
OpenHands + Claude-Opus-4.5   & $-4.6$ & \rankdn{1} & -17.1\% \\
All 37 ($\rho\,0.959$, 59\% unchanged) & $-2.0$ & -- & -24.6\% \\
\textit{TerminalBench (threshold = 0.85): no same scaffold in training} & & & \\
Terminus-2 + Claude-Opus-4.5  & $-0.4$ & 0 & -1.8\% \\
Claude-Code + Claude-Opus-4.5 & $-2.8$ & \rankdn{2} & -20.9\% \\
OpenHands + Claude-Opus-4.5   & $-1.4$ & \rankdn{1} & -16.3\% \\
All 37 ($\rho\,0.994$, 70\% unchanged) & $-0.9$ & -- & -12.7\% \\
\textit{Toolathlon (threshold = 0.90)} & & & \\
Claude-Opus-4.5   & $-1.9$ & 0 & -22.7\% \\
Claude-Sonnet-4.5 & $-2.2$ & 0 & -25.4\% \\
Gemini-3-Pro      & $-1.2$ & 0 & -18.0\% \\
All 22 ($\rho\,0.994$, 70\% unchanged) & $-0.9$ & -- & -23.0\% \\
\end{tblr}
\vspace{-0.2cm}
\end{table}
\subsection{RQ2: Preservation of Per-Agent Rankings}\label{sec:rq2}

To verify whether the relative ordering of agents is preserved under early stopping, we construct an early-stopped leaderboard using the leave-one-agent-out protocol and evaluate it against the full-run ground truth. This analysis utilizes the operating points selected in RQ1 ($0.95$ for SWE-bench Verified, $0.90$ for Toolathlon and the TerminalBench no-same-model split, and $0.85$ for the TerminalBench no-same-scaffold split). Given that the success predictor is reliable exclusively on SWE-bench Verified, we rank agents using the full dual mechanism on SWE-bench, but rely solely on the failure predictor for TerminalBench and Toolathlon. Consequently, the $\Delta$Steps reported for the latter two benchmarks in Table~\ref{tab:rq2} correspond to the failure-only columns in Table~\ref{tab:rq1}. SWE-bench Verified and Toolathlon rank base LLMs under a single, fixed scaffold ($16$ and $22$ agents, respectively), whereas TerminalBench evaluates $37$ distinct scaffold+LLM combinations under each of its two leakage-controlled splits. Ranking fidelity is quantified using Spearman's rank correlation coefficient ($\rho$) and the proportion of agents whose exact ordinal positions are preserved. Table~\ref{tab:rq2} displays the top-3 agents for each setting alongside global leaderboard statistics.

EarlyEval successfully reproduces the full-run agent rankings with high fidelity across all evaluation targets. Spearman's $\rho$ reaches $0.991$ on SWE-bench Verified, $0.994$ on Toolathlon, and $0.994$ on the TerminalBench no-same-scaffold split; the more challenging no-same-model split yields a marginally lower yet highly congruent correlation of $0.959$. The exact-rank fraction corroborates these findings, ranging from $59\%$ on the no-same-model split up to $81\%$ on SWE-bench Verified, with Toolathlon and the no-same-scaffold split intermediate at approximately $70\%$. These results confirm that heterogeneous leaderboards under strict leakage control remain stable, performing on par with single-scaffold rankings.

\begin{tcolorbox}[size=title]
{\textbf{Answer to RQ2:}} EarlyEval consistently preserves downstream agent rankings, yielding Spearman's $\rho$ between $0.959$ and $0.994$ across all evaluation targets, and exceeding $\rho \geq 0.991$ on single-scaffold leaderboards. Furthermore, between $59\%$ and $81\%$ of the evaluated agents maintain their exact ordinal positions.
\end{tcolorbox}

\begin{table}[t]\centering
\small
\caption{RQ3: Feature ablation study on SWE-bench Verified evaluating the robustness of EarlyEval to the omission of specific feature families and constituent subgroups.}
\label{tab:rq3}
\scriptsize
\begin{tblr}{
  colspec   = {Q[l,wd=3.6cm] *{4}{Q[c,wd=1.25cm]}},
  colsep    = 0pt,
  rowsep    = 1.3pt,
  row{1}    = {font=\bfseries},
  row{2}    = {font=\bfseries},
  hline{1}  = {0.08em},
  hline{2}  = {0.05em},
  hline{3}  = {0.05em},
  hline{7}  = {0.05em},
  hline{13} = {0.05em},
  hline{17} = {0.05em},
  hline{Z}  = {0.08em},
  cell{3}{1}  = {c=5}{l},
  cell{7}{1}  = {c=5}{l},
  cell{13}{1} = {c=5}{l},
  cell{17}{1} = {c=5}{l},
}
Feature set & Coverage & Accuracy & $\Delta$Steps & $\Delta$$|$Pass@1$|$ \\
Full (all features) & 34.8\% & 95.0\% & -26.0\% & 1.1\% \\
\textit{Remove one feature family} & & & & \\
\quad w/o Behavioral & 23.4\% & 94.7\% & -16.4\% & 0.8\% \\
\quad w/o Textual & 35.9\% & 94.5\% & -26.5\% & 1.2\% \\
\quad w/o Reference & 32.1\% & 93.9\% & -24.7\% & 1.2\% \\
\textit{Remove one behavioral group} & & & & \\
\quad w/o Activity counts & 35.3\% & 94.8\% & -26.7\% & 1.2\% \\
\quad w/o Last step & 34.9\% & 94.9\% & -26.2\% & 1.2\% \\
\quad w/o Event timing & 34.5\% & 95.1\% & -25.9\% & 1.1\% \\
\quad w/o Working pattern & 34.9\% & 94.5\% & -26.1\% & 1.3\% \\
\quad w/o Error \& test status & 34.7\% & 95.0\% & -26.1\% & 1.2\% \\
\textit{Remove one textual group} & & & & \\
\quad w/o Task prompt & 35.0\% & 94.5\% & -26.7\% & 1.2\% \\
\quad w/o Action text & 34.8\% & 95.0\% & -25.9\% & 1.1\% \\
\quad w/o Feedback text & 35.3\% & 95.5\% & -26.2\% & 1.1\% \\
\textit{Remove one reference group} & & & & \\
\quad w/o Gold descriptors & 33.5\% & 94.1\% & -25.7\% & 1.3\% \\
\quad w/o Prefix--gold overlap & 34.1\% & 94.8\% & -26.1\% & 1.2\% \\
\end{tblr}
\vspace{-0.2cm}
\end{table}
\subsection{RQ3: Robustness to Feature Availability}

The results in RQ1 demonstrate that EarlyEval retains high evaluation fidelity on the reference-free TerminalBench and Toolathlon benchmarks without access to the Reference-Solution feature family (Table~\ref{tab:rq1}). To elucidate the mechanisms underlying this robustness to missing inputs, we perform a systematic feature ablation study on SWE-bench Verified, the sole benchmark where all three feature families are concurrently available. We omit one feature family, or one constituent group within a family, at a time, retrain both predictors under the identical leave-one-agent-out protocol, and compare the performance against the full-feature baseline ($34.8\%$ coverage, $95.0\%$ accuracy, $26.0\%$ step savings, and $1.1\%$ $\Delta|\text{Pass@1}|$). Table~\ref{tab:rq3} summarizes the coverage, prediction accuracy, step savings, and evaluation fidelity for each variant.

The results show that EarlyEval is highly resilient to the omission of individual feature groups. Within the textual modality, removing either the task prompt or individual process-derived blocks perturbs coverage and accuracy by at most $0.5$ percentage points, leaving the early-stopping dynamics virtually unchanged. Notably, completely discarding the Reference-Solution family only marginally reduces coverage to $32.1\%$ and step savings to $24.7\%$. This directly explains why EarlyEval generalizes effectively to reference-free evaluation environments. 
In contrast, the Behavioral family emerges as the primary driver of early stopping; its complete removal drops coverage to $23.4\%$ and step savings to $16.4\%$, marking the most substantial performance degradation among all configurations. This significance, however, is collectively distributed rather than concentrated: ablating any single behavioral group alters coverage by at most $0.5$ percentage points (and any reference subgroup by at most $1.3$). This indicates that each family carries early-stopping signals that are redundantly encoded across its constituent sub-features—a decoupling property that enables EarlyEval to operate robustly across diverse benchmarks with heterogeneous feature availability.

\begin{tcolorbox}[size=title]
{\textbf{Answer to RQ3:}} EarlyEval's predictive signals are redundantly distributed both across and within individual feature families. Consequently, omitting a specific family or subgroup primarily scales down the early-stopping coverage and subsequent compute savings rather than compromising decision quality.
\end{tcolorbox}

\begin{table}[t]\centering
\small
\caption{RQ4: Architectural backbone ablation on SWE-bench Verified comparing LightGBM against alternative classification models and LLM judge baselines.}
\label{tab:rq4}
\scriptsize
\begin{tblr}{
  colspec   = {Q[l,wd=3.6cm] *{4}{Q[c,wd=1.3cm]}},
  colsep    = 0pt,
  rowsep    = 1.3pt,
  row{1}    = {font=\bfseries},
  row{2}    = {font=\bfseries},
  hline{1}  = {0.08em},
  hline{2}  = {0.05em},
  hline{Z}  = {0.08em},
}
Variant & Coverage & Accuracy & $\Delta$Steps & $\Delta$$|$Pass@1$|$ \\
LightGBM (ours) & 34.8\% & 95.0\% & -26.0\% & 1.1\% \\
Direct MLP & 26.9\% & 87.9\% & -20.0\% & 3.3\% \\
Linear (dense LR) & 9.7\% & 43.8\% & -7.7\% & 5.5\% \\
Linear (TF-IDF LR) & 2.4\% & 79.5\% & -2.0\% & 0.3\% \\
Local LLM judge (Qwen LoRA) & 18.7\% & 90.7\% & -17.9\% & 0.8\% \\
\end{tblr}
\vspace{-0.2cm}
\end{table}

\subsection{RQ4: Architectural Ablation}

Finally, we scrutinize the architecture of the predictor backbone itself. Holding EarlyEval's dual success/failure architecture (Section~\ref{subsec:model}) constant, we ablate the per-head classification backbone by evaluating a Multilayer Perceptron (Direct MLP) and two logistic regression baselines: one optimized over the dense feature vector $\phi$ (Linear, dense LR) and another optimized over the raw TF-IDF text features prior to SVD (Linear, TF-IDF LR). Additionally, we compare these against a LoRA-fine-tuned Qwen-0.5B judge model that directly processes the raw trajectory text (Qwen LoRA). All architectural variants are evaluated on SWE-bench Verified under the identical leave-one-agent-out protocol and calibrated at the $0.95$ dual threshold established in RQ1.

As illustrated in Table~\ref{tab:rq4}, LightGBM defines the Pareto frontier of the accuracy--efficiency trade-off, achieving the highest coverage ($34.8\%$), accuracy ($95.0\%$), and step reduction ($26.0\%$), while maintaining a minimal $\Delta|\text{Pass@1}|$ distortion of $1.1$ percentage points. Both the neural and dense-linear backbones are strictly dominated across all four evaluation axes. Specifically, the Direct MLP terminates fewer execution runs ($26.9\%$ coverage) at a lower accuracy ($87.9\%$), saving only $20.0\%$ of execution steps while inflating metric distortion to $3.3$ points. Meanwhile, the dense logistic regression collapses to $43.8\%$ and introduces the largest fidelity error in the study ($5.5$ points), demonstrating that a linear decision boundary over dense features cannot adequately separate non-linear textual outcomes.

The TF-IDF logistic regression baseline achieves a low distortion of $0.3$ points, but it triggers on only $2.4\%$ of trajectories, resulting in a negligible $2.0\%$ step reduction. Its high accuracy ($79.5\%$) is a trivial consequence of its passivity, as it almost never intervenes. Consequently, neither linear baseline offers a viable cost--fidelity operating point.

The fine-tuned Qwen judge represents the only baseline that is genuinely competitive in terms of fidelity, achieving $90.7\%$ accuracy and a low $0.8$-point metric distortion while halting a reasonable proportion of runs ($18.7\%$ coverage). However, it falls short of LightGBM in efficiency, saving roughly half as many execution steps ($17.9\%$ step reduction versus $26.0\%$). Furthermore, its marginally lower distortion is an artifact of its lower coverage—less intervention inherently leaves less room for metric distortion. Crucially, the LLM judge requires a costly model forward pass at every trajectory step, meaning its runtime inference overhead directly offsets the computational savings that early stopping is intended to achieve. In contrast, LightGBM evaluates the same feature vector in sub-millisecond CPU time.

\begin{tcolorbox}[size=title]
{\textbf{Answer to RQ4:}} Only the LightGBM backbone simultaneously achieves high predictive accuracy ($95.0\%$) and high operational activity ($34.8\%$ coverage, $26.0\%$ step savings)..
\end{tcolorbox}

\section{Threats to validity} \label{sec:threats}

\noindent \textbf{Internal validity}
The primary threat to internal validity lies in the potential data leakage during the training.  We strictly mitigated this by implementing a task-partitioned leave-one-agent-out protocol. All prefixes originating from a specific agent trajectory were strictly isolated on one side of the split.

\noindent \textbf{External validity}
External validity concerns the generalizability of EarlyEval to completely unseen agent architectures and novel benchmarks. We addressed this by testing EarlyEval on three structurally heterogeneous benchmark suites (SWE-bench Verified, TerminalBench, and Toolathlon).
Furthermore, our evaluation adopted a realistic deployment scenario by holding out one complete agent configuration at a time during evaluation. Since EarlyEval relies significantly on redundant behavioral and textual features, benchmarks lacking ground-truth human reference patches can still seamlessly utilize the framework by disabling the reference-solution feature family with minimal performance degradation.

\noindent \textbf{Construct Validity}
Construct validity evaluates whether our proxy metrics ($\Delta \text{Steps}$, $\Delta \text{Token}_{\text{in}}$, $\Delta \text{Token}_{\text{out}}$) faithfully capture actual resource reductions without distorting the benchmark's downstream performance assessments. While token bills do not always scale linearly with step counts due to compounding context windows, our empirical data demonstrates that input-token savings systematically exceed executed step reductions. This confirms that our framework effectively targets and truncates the most bloated, computationally expensive trailing steps of agent trajectories. 
\section{Related Works}

\noindent \textbf{Efficient Benchmarking}
Efficient benchmarking aims to reduce evaluation cost while preserving leaderboard conclusions.
Early work showed that large benchmark suites contain substantial redundancy, so reliable rankings can be recovered from far fewer queries: Anchor Points selects representative examples to approximate benchmark outcomes~\cite{Viv23}, Efficient Benchmarking studies coarse-to-fine budget allocation~\cite{Per23}, and tinyBenchmarks and related distillation methods build small proxy test sets to preserve score estimates and rank correlations~\cite{Pol24}
Adaptive testing makes this procedure more responsive by modeling item difficulty and discrimination to select maximally informative examples~\cite{Yan23,Li25c}, and Fluid Language Model Benchmarking picks items based on a model's current ability, improving efficiency while reducing saturation and variance~\cite{Hof25}.
A parallel line treats efficient evaluation as subset selection and statistical inference: building informative evaluation subsets~\cite{Bea25,Smo26}, recovering reliable conclusions from partial observations~\cite{Wu26c,Tol26}, and estimating abilities under a limited budget through active selection or cached responses~\cite{Kru26,Liu26e,Pur26,Hel26}.
These ideas have recently reached agent benchmarks, where filtering to mid-difficulty tasks cuts cost while preserving leaderboard fidelity under scaffold and temporal shift~\cite{Ndz26}, and task-level performance prediction is studied from a latent-measurement perspective~\cite{Ge26}. Rather than selecting fewer tasks or examples, we introduce a new axis of efficiency within each task, stopping an agent trajectory early and using partial trajectories so far to estimate the final evaluation result.

\noindent \textbf{Early Stopping for Agents}
Early stopping refers to terminating a process before its predetermined endpoint once continuing is judged unlikely to yield further benefit. The underlying principle predates LLM agents. In automated program repair, for instance, earlier work asked whether a failing test should be delegated to an expensive repair procedure at all, framing the decision around whether continued effort is likely to pay off~\cite{le2015should}. The idea has recently been adapted to LLM agents, whose multi-step interaction loops can consume large numbers of tokens, time, and energy before producing a final answer~\cite{DBLP:conf/kbse/ShiQZSG25,DBLP:journals/corr/abs-2601-16746,DBLP:journals/corr/abs-2607-18213,DBLP:journals/corr/abs-2602-01785}. These include intrinsic exit instructions or task-completion verification~\cite{Lu25}, predictive execution signals like token log-probabilities and uncertainty estimates~\cite{Pha26,DBLP:journals/corr/abs-2606-09577}, remaining budget estimations~\cite{Lin26}, and information sufficiency or learned value-based stopping policies~\cite{Liu25,Par25,Fan26}. What unifies these approaches is their reliance on runtime self-observation to preserve the individual agent's success. Our setting departs from this on both counts. We study early stopping during \emph{benchmark evaluation}, which exposes a different class of signals that deployment-time methods cannot assume, such as reference answers and historical evaluation traces from other agents. Rather than preserving an individual agent's success, our objective is to stop a rollout once additional interaction is unlikely to change its final judged outcome.

\section{Discussion}
\noindent\textbf{Applicability assumption.}
EarlyEval requires a pool of completed, outcome-labeled trajectories on the target benchmark to train its predictors. This is satisfied for the established benchmarks that dominate agent evaluation, which ship with baseline runs and accrue leaderboard submissions. All pools used here were assembled from such public sources. The framework therefore targets the common case of repeatedly evaluating evolving agents against a stable benchmark, where per-pass cost compounds across iterations, rather than the first-ever evaluation of a brand-new benchmark with no prior runs, for which no early-prediction signal yet exists. As a benchmark accumulates runs, the predictors can be refreshed at negligible marginal cost.

\noindent\textbf{Intended use and scope.}
EarlyEval is designed to provide a cheap, high-fidelity signal during iterative development, the regime where a team re-evaluates an evolving agent many times and the dominant concern is relative comparison rather than the exact resolve rate.
It is not intended to replace full execution for producing the canonical, citable benchmark scores: because early stopping introduces a small (${\sim}$1--2~pp) systematic deviation in measured resolve rates, final leaderboard entries and headline claims should still be obtained by running agents to completion. Our fidelity analysis (Section~\ref{sec:rq2}) bounds this distortion precisely, so that practitioners can decide when an early-stopped estimate suffices and when a full run is warranted.

\section{Conclusion}
In this work, we introduced \textbf{EarlyEval}, an effective paradigm designed to lower the financial and computational barriers of LLM agent benchmarking through early outcome prediction. Driven by the insight that an agent's ultimate success or failure is frequently discernible from its intermediate behaviors long before natural completion, EarlyEval implements a highly efficient inference workflow powered by LightGBM tree ensembles. 
Extensive leave-one-agent-out evaluations across three diverse agentic benchmarks, SWE-bench Verified, TerminalBench, and Toolathlon, demonstrate that EarlyEval curtails environment execution steps by 13\% to 26\% and slashes compounding context input tokens by up to 44.1\%. 

\section*{Data Availability Statement}
The code and related experimental data in this paper are accessible on \url{https://github.com/inphotoo/earlyeval}

% ==========================================
% References
% ==========================================
% \newpage
\bibliographystyle{IEEEtran}
\bibliography{main}

\end{document}